\documentclass[11pt]{article}       
\usepackage{geometry}               
\usepackage{tikz}
\usetikzlibrary{shapes.geometric, arrows}

\usepackage{graphicx}
\usepackage{caption}
\usepackage{subcaption}
\usepackage{comment}
\usepackage{amsmath} 
\usepackage{amssymb}  
\usepackage{amsthm}  
\usepackage{bm}
\usepackage{lipsum}
\usepackage{color}
\usepackage{multirow}
\usepackage{array}
\usepackage{cite}
\usepackage{hyperref}

\usepackage{algorithm}
\usepackage{algpseudocode}

\numberwithin{equation}{section}

\title{Censored Sampling for Topology Design: Guiding Diffusion with Human Preferences}

\author{
Euihyun Kim\footnote{Equal contribution, Dept. of Mechanical System Design Engineering, SeoulTech}
\and
\and
Keun Park\footnote{Equal contribution, Dept. of Mechanical System Design Engineering, SeoulTech}
\and
Yeoneung Kim\footnote{Corresponding author, Dept. of Applied Artificial Intelligence, SeoulTech, yeoneung@seoultech.ac.kr}
}
\date{}

\begin{document}
\maketitle

\pagestyle{myheadings}
\thispagestyle{plain}

\begin{abstract}
Recent advances in denoising diffusion models have enabled rapid generation of optimized structures for topology optimization. However, these models often rely on surrogate predictors to enforce physical constraints, which may fail to capture subtle yet critical design flaws such as floating components or boundary discontinuities that are obvious to human experts. In this work, we propose a novel human-in-the-loop diffusion framework that steers the generative process using a lightweight reward model trained on minimal human feedback. Inspired by preference alignment techniques in generative modeling, our method learns to suppress unrealistic outputs by modulating the reverse diffusion trajectory using gradients of human-aligned rewards. Specifically, we collect binary human evaluations of generated topologies and train classifiers to detect floating material and boundary violations. These reward models are then integrated into the sampling loop of a pre-trained diffusion generator, guiding it to produce designs that are not only structurally performant but also physically plausible and manufacturable. Our approach is modular and requires no retraining of the diffusion model. Preliminary results show substantial reductions in failure modes and improved design realism across diverse test conditions. This work bridges the gap between automated design generation and expert judgment, offering a scalable solution to trustworthy generative design.
\end{abstract}

\section{Introduction}
Designing optimal structural topologies under load and support constraints is a longstanding challenge in engineering, with implications for material efficiency and manufacturability. While structural topology optimization (STO) offers a principled framework for this task, classical methods such as SIMP remain computationally expensive and often yield designs requiring post-processing to meet manufacturing criteria.

 While classical approaches rely on physics-based optimization (e.g., SIMP), recent work has explored machine learning based methods that bypass iterative simulation. These include supervised learning frameworks that directly map physical fields to optimized topologies~\cite{yu2019deep, nie2021topologygan, rade2023deep}, often using CNNs or implicit neural representations trained on large datasets. For example, NTopo~\cite{zehnder2021ntopo} employs implicit neural representations to enable mesh-free optimization with continuous design parameterization, while SOLO~\cite{deng2022self} achieves rapid surrogate-guided refinement by integrating learned models with FEM feedback.
 Another line of work formulates topology design as a sequential decision-making process, using reinforcement learning (RL) agents trained to construct high-performance structures under physical constraints~\cite{choi2025deep}. While these ML-based approaches offer improved scalability and flexibility, they often depend on large-scale synthetic data or simulation-based rewards, and still struggle to guarantee structural feasibility, particularly in cases involving boundary connectivity or geometric singularities. 
 
 More recently, generative diffusion models have emerged as a powerful alternative for structure generation. TopoDiff~\cite{maze2023diffusion}, for instance, formulates topology synthesis as a conditional denoising process guided by surrogate predictors for compliance and connectivity. However, the reliance on fixed surrogates still leads to failure cases that escape automated detection, such as thin ligaments or detached supports. In practical engineering applications, such structural defects can lead to critical failure or render a design unprintable in additive manufacturing. For example, partial boundary detachment or floating islands may escape surrogate penalties but cause real-world fabrication issues, compromising structural integrity or function. Addressing such failures is essential for deploying generative models in safety-critical or high-precision domains.

In this work, we propose a complementary approach that combines the expressivity of diffusion models with the flexibility of reward-based guidance without the need for reward engineering. Inspired by recent advances in human preference alignment for generative models~\cite{yoon2023censored}, we train lightweight reward models from minimal human feedback and integrate them into the diffusion sampling process. These learned rewards steer generation toward physically plausible and manufacturable designs that reflect expert judgment.




\section{Related work}
\subsection{Structural topology optimization and data-driven design}
STO formulates the task of distributing material within a fixed design domain to achieve optimal structural performance under given loads and constraints. Classical methods, such as the Solid Isotropic Material with Penalization (SIMP)~\cite{bendsoe2013topology,bendsoe1988generating} and Moving Morphable Components (MMC)~\cite{liu2018efficient,bai2020hollow}, solve this problem via iterative PDE-constrained optimization, typically requiring repeated finite element analysis (FEA). While mathematically rigorous, these approaches are computationally expensive, sensitive to initialization, and often yield intermediate designs that violate manufacturability criteria such as connectivity or minimal feature size.

Early methods~\cite{yu2019deep,ulu2016data,rade2023deep} employed supervised learning to directly map loading and boundary conditions to optimized topologies. These models are typically trained on synthetic datasets generated from classical solvers (e.g., SIMP), enabling fast inference but often lacking robustness, diversity, or constraint enforcement. To overcome these limitations, subsequent works explored conditional generative models that learn to produce diverse and constraint-aware structures given physical field inputs. Conditional generative adversarial networks (cGANs), such as TopologyGAN~\cite{nie2021topologygan}, introduce generative capabilities by conditioning on physical field data (e.g., stress maps), improving visual quality and diversity. However, these models still struggle to enforce hard constraints and often produce invalid or unrealistic outputs due to training instability and lack of explicit physical grounding.

These developments highlight the need for generative frameworks that combine the expressive power of deep learning with mechanisms for enforcing structural feasibility, either through physics-informed constraints or through feedback from human experts.

\subsection{Diffusion models for generative design}

Denoising diffusion probabilistic models (DDPMs) have emerged as a powerful alternative to GANs for generative design due to their stability, expressivity, and flexibility for conditional generation. In particular, TopoDiff~\cite{maze2023diffusion} introduced a conditional DDPM for topology optimization, trained on optimized structures and guided at inference time by surrogate models: a compliance regressor and a floating-material classifier. These surrogates provide gradient-based guidance during the reverse diffusion process to steer generations toward physically feasible and high-performance outputs.

Subsequent works further explored the integration of physics into the diffusion framework. Diffusion Optimization Models (DOM)~\cite{giannone2023aligning} aligned the denoising trajectory with optimization updates, improving sample efficiency. Physics-informed diffusion models~\cite{bastek2024physics} directly incorporated the residuals of PDE constraints into the training objective, allowing the model to learn the manifold of physically admissible solutions.

However, these approaches still rely on approximate or handcrafted surrogates to enforce feasibility. This raises concerns when surrogates fail to detect subtle or high-level design flaws, such as disconnected ligaments or geometric artifacts that are readily apparent to human experts but difficult to encode explicitly. This motivates the need for incorporating human judgment more directly into the generative loop.

\subsection{Human preference alignment in generative modeling}
Human-in-the-loop feedback has become a central theme in aligning generative models with subjective or hard-to-specify criteria. In natural language processing, reinforcement learning from human feedback (RLHF) has enabled large language models to better reflect human intent~\cite{christiano2017deep,ouyang2022training,stiennon2020learning,ziegler2019fine,lee2023aligning}. In image generation, classifier-guided diffusion~\cite{dhariwal2021diffusion} and its extensions have allowed users to guide sample quality or content through auxiliary classifiers.

More recently, censored sampling~\cite{yoon2023censored} introduced a lightweight and efficient method for preference-guided diffusion. This technique trains a small reward model on binary human feedback (e.g., desirable vs undesirable samples), and integrates it into the diffusion sampling loop via a mean-shift update that suppresses undesired outputs. Crucially, this method requires no retraining of the generative model and operates solely at inference time, making it easy to integrate into existing systems.

\subsection{Our contribution in context}
While TopoDiff and its successors have advanced generative topology optimization using physical surrogates, they remain limited in addressing nuanced design failures beyond the surrogate’s expressivity. In contrast, our work is the first to apply censored sampling with human feedback to topology design. By training lightweight classifiers on binary human evaluations of generated designs (e.g., boundary condition violations, floating structures), we guide the diffusion process toward structurally realistic and manufacturable outputs.

Our method represents a synthesis of two threads of research: (1) surrogate-guided diffusion for physics-aware design, and (2) human-in-the-loop alignment for controllable generation. The resulting framework enhances the reliability and trustworthiness of generative STO pipelines, and opens new directions for integrating expert knowledge into design automation.

To summarize, we propose the framework of Human-in-the-loop diffusion for topology optimization:
    \begin{itemize}
        \item Introduce a human-guided reward model into the generation process of topology.
        \item Designs identified as unrealistic by human feedback receive low reward and are suppressed.
        \item Inspired by human preference alignment in AI: adapted to structural design.
        \item Enables generation of diverse, high-performance, and manufacturable topologies.
    \end{itemize}

\section{Background: guided diffusion and TopoDiff}
\subsection{Conditional diffusion models for design generation}
Denoising diffusion probabilistic models (DDPMs)~\cite{ho2020denoising} define a generative process by learning to reverse a Markovian forward process that gradually adds Gaussian noise to data. Given a data point $x_0 \sim p_{\text{data}}(x)$, the forward process defines a sequence $\{x_t\}_{t=1}^T$ via
\[
q(x_t \mid x_{t-1}) = \mathcal{N}(x_t; \sqrt{1 - \beta_t} \, x_{t-1}, \beta_t , I),
\]
with a variance schedule $\{\beta_t\}_{t=1}^T$. The reverse process is approximated by a neural network $\epsilon_\theta(x_t, t)$ trained to predict the noise in the forward process.

In a conditional generation setting, additional problem-specific context $c$ (e.g., load fields, boundary masks) is provided at each step. The model learns to denoise $x_t$ into a structure $x_0$ consistent with both training data and conditioning input:
\[
x_{t-1} = \mu_\theta(x_t, t, c) + \Sigma_t^{1/2} \cdot \epsilon, \quad \epsilon \sim \mathcal{N}(0, I),
\]
where the $\mathcal{N}(0,I)$ denotes the standard normal distribution.

\subsection{TopoDiff: Diffusion-based topology optimization}
Classical topology optimization methods such as SIMP are computationally expensive and often require heuristic filters to enforce manufacturability. Recent deep generative models (e.g., GANs) offer faster design synthesis but struggle to directly incorporate physics-based objectives such as compliance or structural feasibility. To bridge this gap, \textbf{TopoDiff} formulates topology generation as a conditional image denoising process guided by physical performance metrics. By leveraging the sampling flexibility of diffusion models, TopoDiff enables high-quality structure generation while allowing direct guidance toward compliance and connectivity.

TopoDiff~\cite{maze2023diffusion} is a conditional DDPM framework tailored to structural topology optimization. The goal is to synthesize binary material layouts $x \in \{0,1\}^{H \times W}$ under given loading and support conditions. Each conditioning input $c \in \mathbb{R}^{6 \times H \times W}$ encodes six spatial fields: volume fraction, von Mises stress, strain energy density, and external loads in the $x$ and $y$ directions. The physical fields are computed using FEA and implicitly capture the effects of boundary supports and applied loads.

The model uses a UNet-based architecture with attention, trained on thousands of optimized 2D topologies generated via SIMP-based solvers. The DDPM loss is formulated as:
\[
\mathcal{L}_{\text{DDPM}} = \mathbb{E}_{x_0, t, \epsilon} \left[ \left\| \epsilon - \epsilon_\theta\left(\sqrt{\bar{\alpha}_t} x_0 + \sqrt{1 - \bar{\alpha}_t} \epsilon, t, c \right) \right\|^2 \right].
\]

To improve design feasibility and performance, TopoDiff applies surrogate-based guidance during sampling. Two neural surrogates are trained:

\begin{itemize}
    \item \textbf{Compliance regressor} $c_\phi(x, c)$: predicts structural compliance, guiding toward stiffer designs.
    \item \textbf{Floating-part classifier} $p_\gamma(\text{feasible} \mid x, c)$: detects disconnected or unsupported material regions.
\end{itemize}

The reverse mean is then updated via:
\[
\mu \leftarrow \mu 
- \eta_c \Sigma\nabla_x   c_\phi(x_t, c) 
+ \eta_f \Sigma\nabla_x \Sigma \log p_\gamma(x_t, c),
\]
where $\eta_c$ and $\eta_f$ are regressor and classifier gradient scale hyperparameters. This is analogous to classifier guidance in image diffusion~\cite{dhariwal2021diffusion}, and improves compliance and connectivity without retraining the diffusion model.

During sampling, the reverse diffusion process predicts a denoised image at each step by computing the mean of a conditional Gaussian. Our method perturbs this predicted mean in the direction of decreasing compliance and improving connectivity, effectively steering the generation toward physically realistic and high-performance structures. This allows performance-driven guidance to be injected at inference time without modifying the trained diffusion model.

\subsection{Limitations of surrogate-based guidance}
\begin{figure}[H]
    \centering
    \includegraphics[width=1\linewidth]{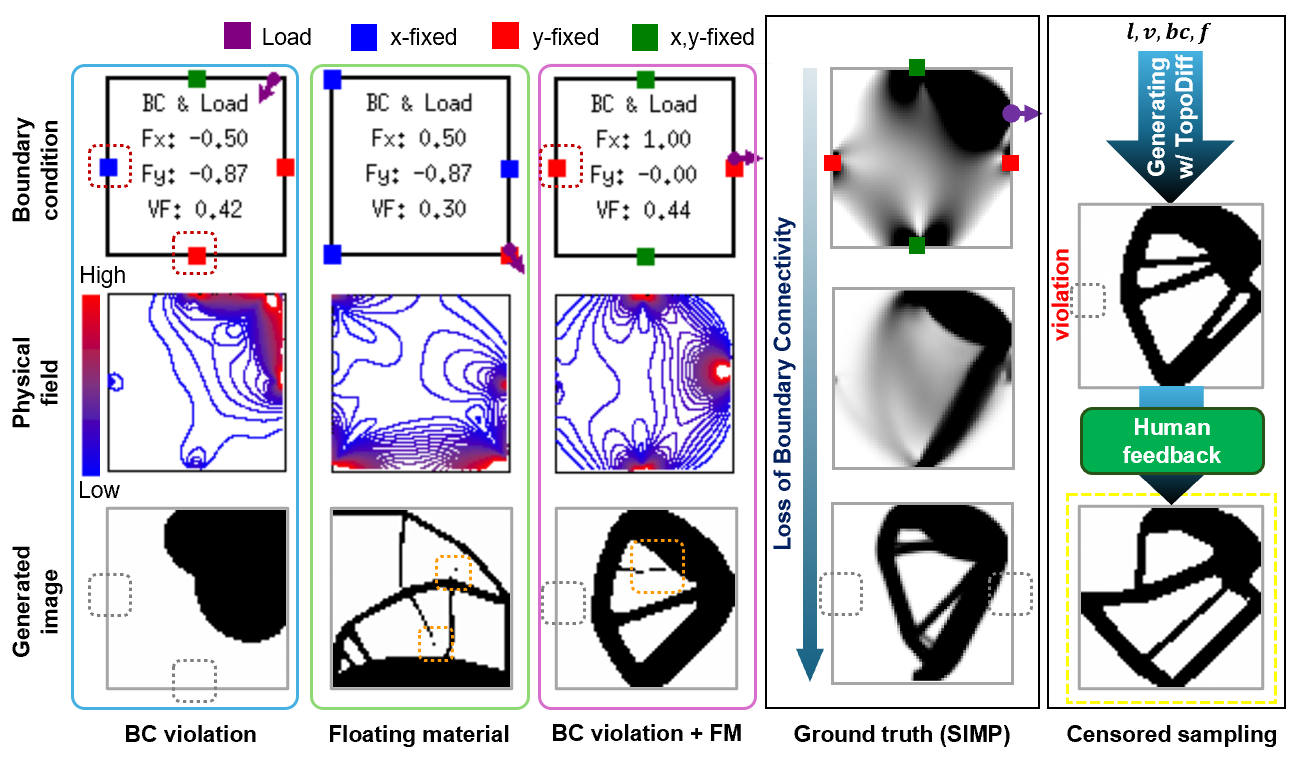}
    \caption{
\textbf{Left}: Representative failure cases illustrating the limitations of surrogate-based guidance in TopoDiff. 
Each example corresponds to a distinct boundary condition and load configuration (top row), with associated physical field visualization (middle row) and generated topology (bottom row).
Despite physically plausible conditions, the generated structures violate basic design principles such as boundary connectivity or support attachment. 
\textbf{Right}: Human feedback identifies such violations and enables improved generation through reward-guided diffusion.
Dashed boxes highlight regions where connectivity is broken or material is detached from supports.
}
    \label{fig:limitation}
\end{figure}
While surrogate models have enabled constraint-aware generative design in diffusion frameworks like TopoDiff, they remain limited in generalization and sensitivity. Most methods rely on compact signals, such as compliance values or binary feasibility labels, trained on synthetic data. As a result, they often fail to penalize structural flaws that are visually evident but difficult to parameterize.

Figure~\ref{fig:limitation} showcases failure cases that emerge despite correct boundary and loading conditions. In each case, the generated topology violates fundamental design expectations such as failing to connect to fixed supports or introducing unsupported material regions. These issues typically arise when surrogate signals (e.g., compliance, floating material classifiers) overlook fine-grained geometric errors or context-specific failures.

Although such artifacts may not trigger large compliance penalties, they are immediately judged as unrealistic or unmanufacturable by human evaluators. This discrepancy highlights a key limitation of purely signal-driven guidance and motivates the integration of human-aligned feedback into the generative process.

\section{Proposed method: Human-guided censored sampling}
\label{sec:method}

We enhance the TopoDiff framework by integrating human feedback into the generative sampling process via reward-guided censored sampling. Our approach introduces lightweight reward models trained from minimal binary feedback to steer the reverse diffusion process toward realistic and manufacturable topologies.

\subsection{Human feedback and reward learning}

We begin by collecting binary labels from human experts on a set of generated topologies. Each design $x$ is labeled along two categories: 
\begin{itemize}
    \item \textbf{Boundary condition (BC) validity} $y^{\text{bc}} \in \{0,1\}$:  does the structure maintain boundary integrity?
    \item \textbf{Floating material (FM)} $y^{\text{fm}} \in \{0,1\}$: does the structure contain disconnected or unsupported components?
\end{itemize}
We set $y^{\text{bc}}=1$ (resp.\ $0$) when boundary conditions are satisfied (resp.\ violated), and $y^{\text{fm}}=1$ (resp.\ $0$) when no floating material is present (resp.\ present). The process is conducted via our customized GUI. 

\begin{figure}[ht]
    \centering
    \includegraphics[width=1\linewidth]{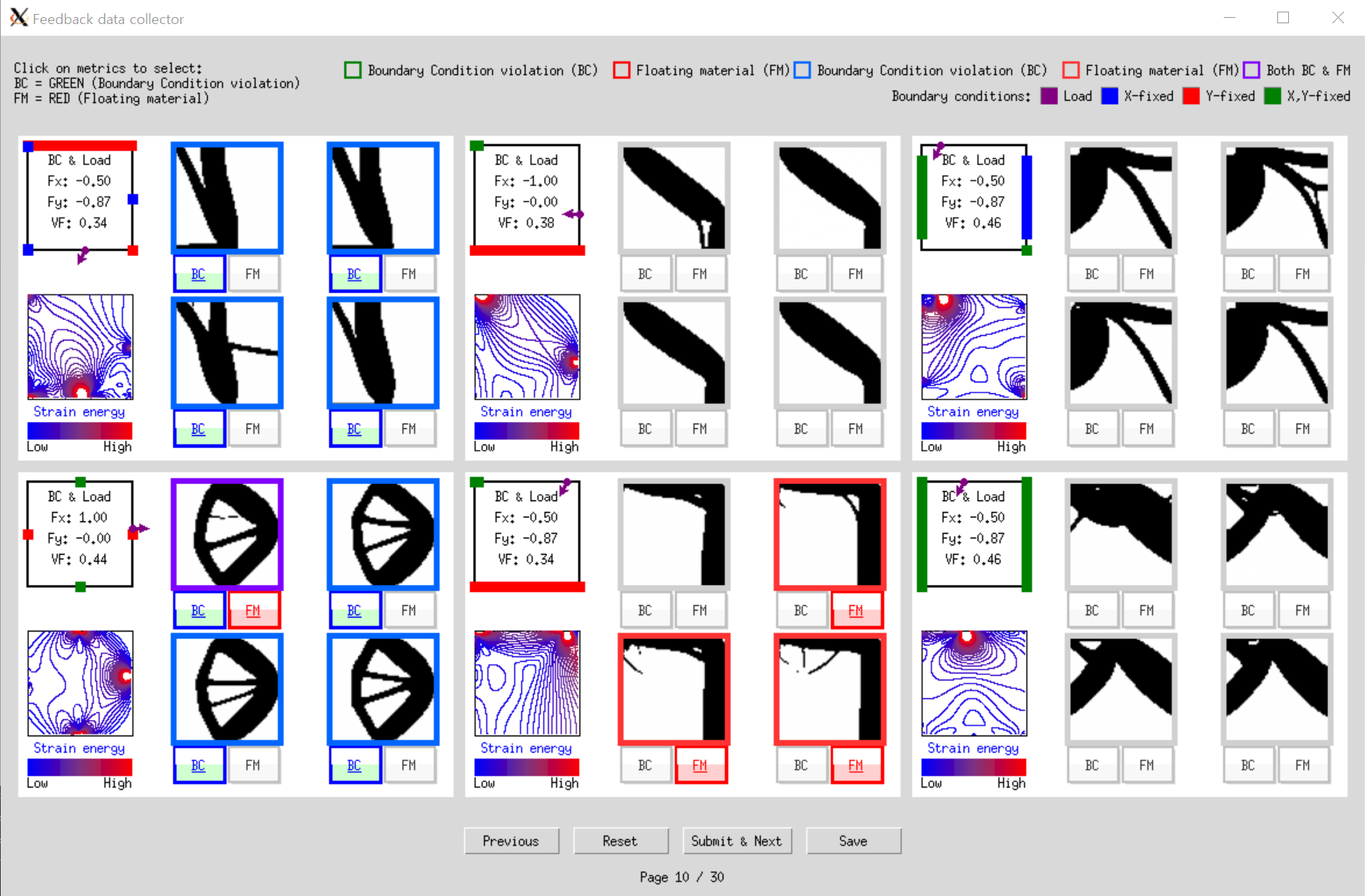}
    \caption{GUI for collecting binary human feedback on generated topologies. Each sample includes boundary and load conditions, physical fields, and the generated structure. Annotators label each sample as exhibiting a boundary condition violation, floating material, or both, using checkboxes based on structural plausibility.}
    \label{fig:feedback-ui}
\end{figure}

\paragraph{Feedback interface.}
To facilitate rapid feedback collection, we implement a minimal GUI designed for binary annotation of generated topologies. As shown in Figure~\ref{fig:feedback-ui}, candidate structures generated under the same loading and boundary condition are displayed in a grid layout. For each sample, annotators indicate whether it is \emph{valid} or \emph{invalid} based on realism and manufacturability criteria (e.g., boundary detachment, disconnected parts), by clicking the BC violation or FM buttons below the image. Each annotation session takes less than two minutes and typically yields over 20 labeled samples, enabling efficient training of reward models from limited supervision.

\paragraph{Training of reward model.}
Once labels are collected, we facilitate them to train two reward models:
\begin{itemize}
    \item $R_{\psi_{\text{bc}}}(x, c, t)$ estimates the probability that $x$ satisfies boundary conditions.
    \item $R_{\psi_{\text{fm}}}(x, t)$ estimates the absence of floating structures.
\end{itemize}
Both are trained using binary cross-entropy loss on noisy intermediate samples $x_t$ generated during the forward diffusion process.

Each reward model $R_\psi(x_t, t)$ is implemented as a shallow convolutional classifier trained to predict binary human labels (valid or invalid) on noised intermediate states $x_t$. For each failure type (e.g., BC violation or floating material), we collect a few dozen labeled examples and add noise consistent with the diffusion forward process to simulate $x_t$ at various timesteps.

The model is trained using binary cross-entropy loss:
\[
\begin{split}
&\mathcal{L}_\text{reward}(\psi) \\
&= - \mathbb{E}_{(x_t, y)} \left[ y \log R_\psi(x_t, t) + (1 - y) \log (1 - R_\psi(x_t, t)) \right],
\end{split}
\]
where $y \in \{0,1\}$ is the human label. To improve generalization from few labels, we apply data augmentation (e.g., random flipping, jittering) and train the model jointly across multiple timesteps $t$.

Despite the small size of the labeled dataset (often fewer than 100 samples), we find that the reward model generalizes well within the structure space of TopoDiff, effectively capturing common human-flagged failure patterns.

To incorporate these reward signals into the sampling process, we modify the reverse-time mean $\mu_\theta(x_t, t)$ of the DDPM using gradient ascent on the log-reward. For a single reward model $R_\psi$, the updated mean becomes:
\begin{equation*}
    \mu_t' = \mu_\theta(x_t, t) + \lambda \Sigma_t \nabla_{x_t} \log R_\psi(x_t, t),
\end{equation*}
where $\lambda > 0$ is a guidance strength coefficient and $\Sigma_t$ is the variance at step $t$.

When multiple reward models are used, such as for both BC and FM constraints, we combine their gradients:
\begin{align*}
    &\mu_t \leftarrow \mu_t 
    \\
    &\qquad+ \lambda_{\text{bc}} \Sigma_t \nabla_{x_t} \log R_{\psi_{\text{bc}}}(x_t, c, t)
    \\
    &\qquad+ \lambda_{\text{fm}} \Sigma_t \nabla_{x_t} \log R_{\psi_{\text{fm}}}(x_t, t).
\end{align*}

To ensure stability, we apply guidance only within a limited time window $t < \text{MLN}_\phi$, where early timesteps contain sufficient semantic structure. Here, $\text{MLN}_\phi$ denotes a cutoff timestep MLN(maximum level of noise) specific to each reward model $\phi$, beyond which the samples become too structured for stable guidance. We further enhance gradient sharpness by scaling the reward:
\begin{equation*}
    \nabla \log R(x)^\alpha = \alpha \nabla \log R(x), \quad \alpha > 1.
\end{equation*}

In our implementation, we adopt the same hyperparameter settings as TopoDiff. The boundary condition (BC) reward is merged with the compliance regressor and applied consistently throughout the entire denoising process. In contrast, the floating material (FM) reward combined with the connectivity classifier is selectively activated only during the later, low-noise timesteps. This design choice reflects the observation that FM predictions are unreliable under high noise; thus, guidance is deferred until the denoising trajectory reaches semantically meaningful regions.

\subsection{Sampling process}
We now summarize the modified sampling procedure incorporating reward-guided updates. Given a pretrained conditional diffusion model with denoising network $\epsilon_\theta(x_t, t)$, the generation process proceeds as follows at each reverse diffusion step:

\begin{enumerate}
    \item Predict the noise component: $\epsilon_\theta(x_t, t)$.
    \item Compute the standard reverse mean: $\mu_\theta(x_t, t)$.
    \item Evaluate reward gradients $\nabla \log R_{\psi_k}(x_t, t)$ for each reward model $R_{\psi_k}$ (e.g., BC or FM).
    \item Update the reverse mean via:
    \[
    \mu_t \leftarrow \mu_\theta(x_t, t) + \sum_k \lambda_k \Sigma_t \nabla_{x_t} \log R_{\psi_k}(x_t, t),
    \]
    where $\lambda_k$ is the guidance strength for reward $k$, and $\Sigma_t$ is the variance at step $t$.
    \item Sample the next state: $x_{t-1} \sim \mathcal{N}(\mu_t, \Sigma_t)$.
\end{enumerate}
This modification introduces human-aligned guidance without modifying or retraining the pretrained diffusion model. It is fully compatible with existing conditional diffusion frameworks and can be applied in conjunction with other guidance signals (e.g., physics-based surrogates).

Our method retains the architecture and conditional framework of TopoDiff but replaces fixed surrogate guidance with human-aligned reward gradients. Unlike compliance or connectivity surrogates trained on synthetic labels, our reward models reflect expert evaluations of nuanced failure modes. This allows us to suppress artifacts that are easily overlooked by automated predictors, such as extremely thin ligaments, partial boundary violations, or impractical geometries, thereby bridging the gap between physics-aware generation and manufacturable design.

\section{Experiments}
\label{sec:experiments}
We evaluate our method using representative failure cases from the TopoDiff model~\cite{maze2023diffusion}. Instead of evaluating over large-scale randomly sampled datasets, which would require exhaustive human labeling, we focus on a small set of human identified failure modes. Each case illustrates a structural defect commonly overlooked by surrogate models but obvious to human experts.

For each selected case, we run the base TopoDiff model and apply our reward-guided sampling with either $R_{\psi_{\text{bc}}}$, $R_{\psi_{\text{fm}}}$, or both. The goal is to qualitatively assess whether human-guided diffusion can suppress the undesirable features while preserving structural integrity.

\subsection{Case Study A: Fixing boundary condition violations}

\begin{figure}[H]
    \centering
    \includegraphics[width=1\linewidth]{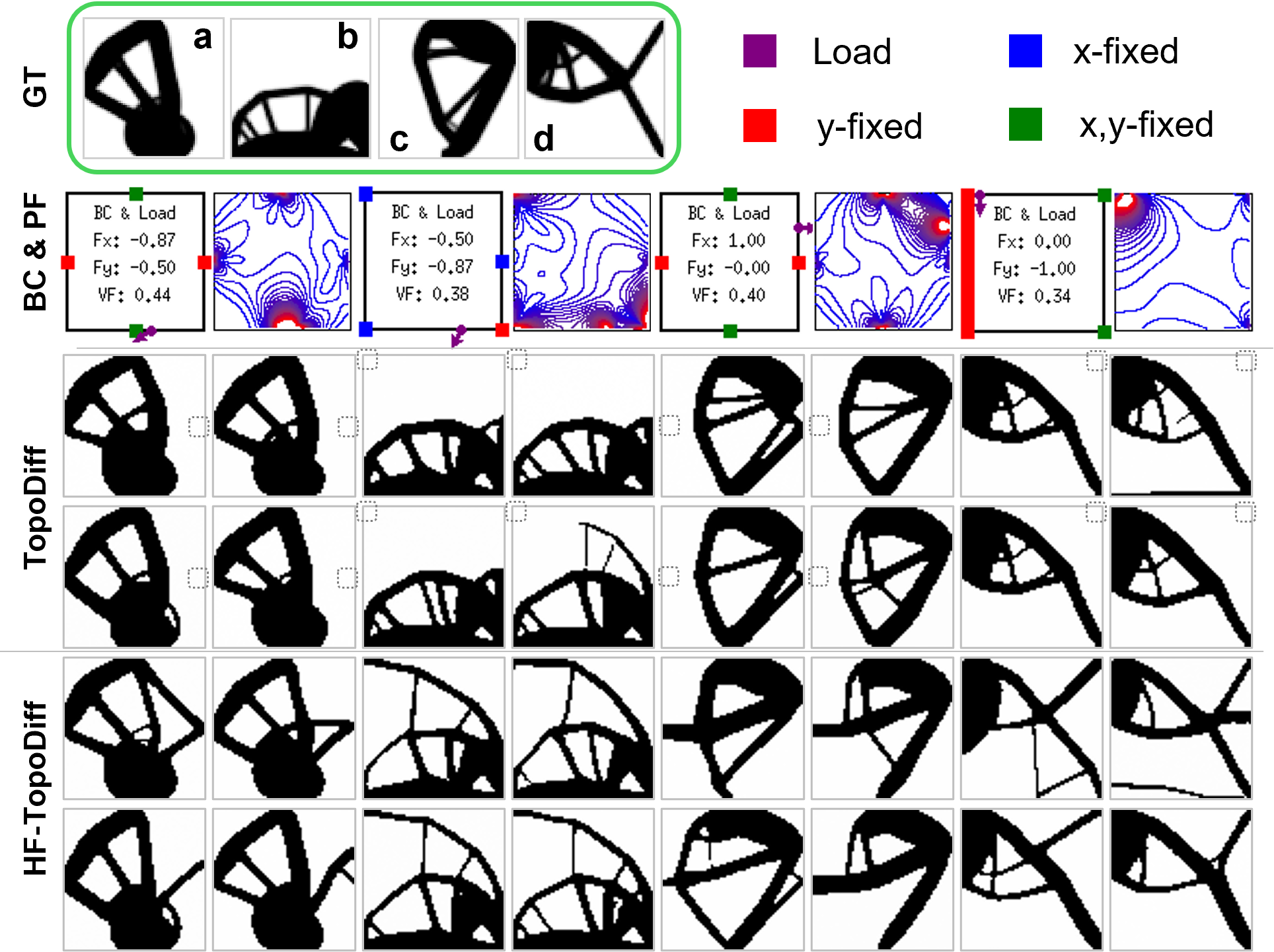}
    \caption{
    Case study of boundary condition violations and their correction using human-guided reward sampling. 
    Each row corresponds to a unique boundary condition and load configuration (shown on the left). 
    The first column shows the TopoDiff baseline output, which exhibits partial or complete detachment from the support region. 
    The next columns show samples generated using our method with increasing guidance strengths $\lambda_{\text{bc}}$. 
    As the reward guidance is strengthened, the generated designs exhibit improved attachment to the prescribed boundaries while preserving structural diversity.
    }
    \label{fig:case-bc}
\end{figure}

We select several representative design tasks where the TopoDiff model produces outputs with broken boundary conditions such as material failing to connect to a fixed support or load-bearing edge. These violations are difficult to penalize via compliance alone, but are immediately recognizable as unmanufacturable by human experts.

We apply our method using the boundary reward model $R_{\psi_{\text{bc}}}$ trained on minimal human feedback. As shown in Figure~\ref{fig:case-bc}, each row corresponds to a distinct task. The first column shows the original TopoDiff outputs, which include clear boundary defects. The remaining columns display samples generated by our method under increasing boundary guidance strength $\lambda_{\text{bc}}$.

In all cases, the guided samples restore boundary connectivity and exhibit physically plausible transitions. Notably, the correction occurs without collapsing structural diversity or introducing artifacts. These results demonstrate the effectiveness of targeted reward guidance in correcting subtle failure modes that elude surrogate models.

\subsection{Case study B: Fixing floating material artifacts}
\begin{figure}[H]

    \centering
    \includegraphics[width=1\linewidth]{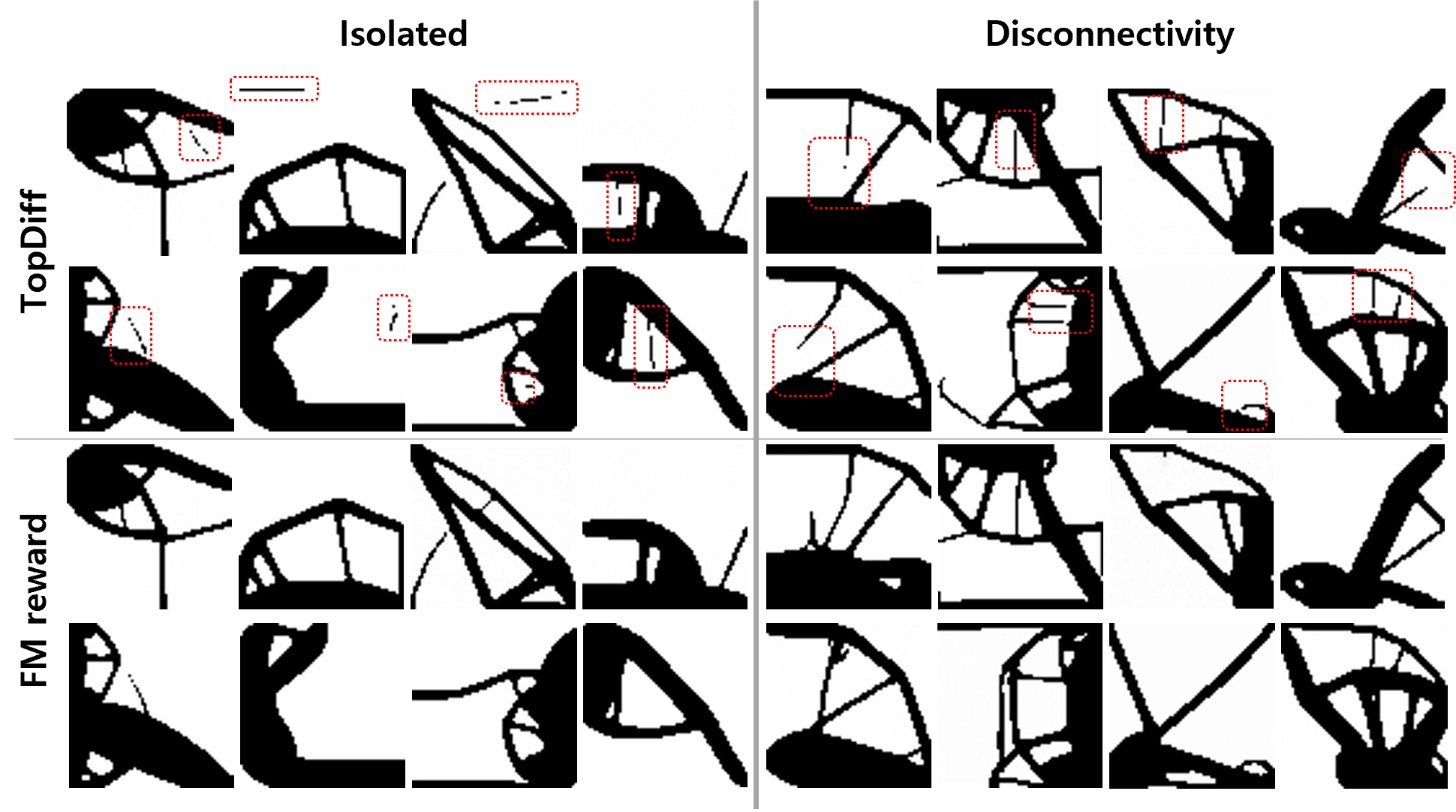}

\caption{
        Floating material failure modes and their correction via human-guided reward sampling.
        {Top two rows:} TopoDiff outputs exhibiting FM artifacts. Red boxes indicate two failure types: (left) isolated material that is entirely disconnected, and (right) disconnectivity where structural segments are weakly attached or broken. 
        {Bottom two rows:} Outputs from our method using the FM reward model. Both isolated and disconnected parts are successfully eliminated or reconnected, producing valid and manufacturable topologies.
    }
    \label{fig:case-fm}

\end{figure}

We categorize floating material (FM) failures into two types: (1) \textbf{Isolated islands}, where material is entirely disconnected from the main structure, and (2) \textbf{Disconnectivity}, where structural components are weakly attached or topologically broken. While both types are structurally infeasible, the latter is more challenging to detect automatically due to subtle geometric inconsistencies.

In this experiment, we collect human labels for both FM types and train a single reward model $R_{\psi_{\text{fm}}}$ that learns to suppress both. Figure~\ref{fig:case-fm} shows comparisons of TopoDiff outputs and our guided samples for representative FM cases. Red boxes highlight violations in the baseline outputs. These include disconnected islands, dangling fragments, and split paths that break load transmission.


As shown in Figure~\ref{fig:case-fm}, FM-guided sampling eliminates isolated fragments and restores connectivity in broken regions, yielding structurally valid outputs that are difficult to obtain with fixed classifiers alone.

\subsection{Case study C: Dual failure and joint reward guidance}

\begin{figure}[H]
    \centering
    \includegraphics[width=1\linewidth]{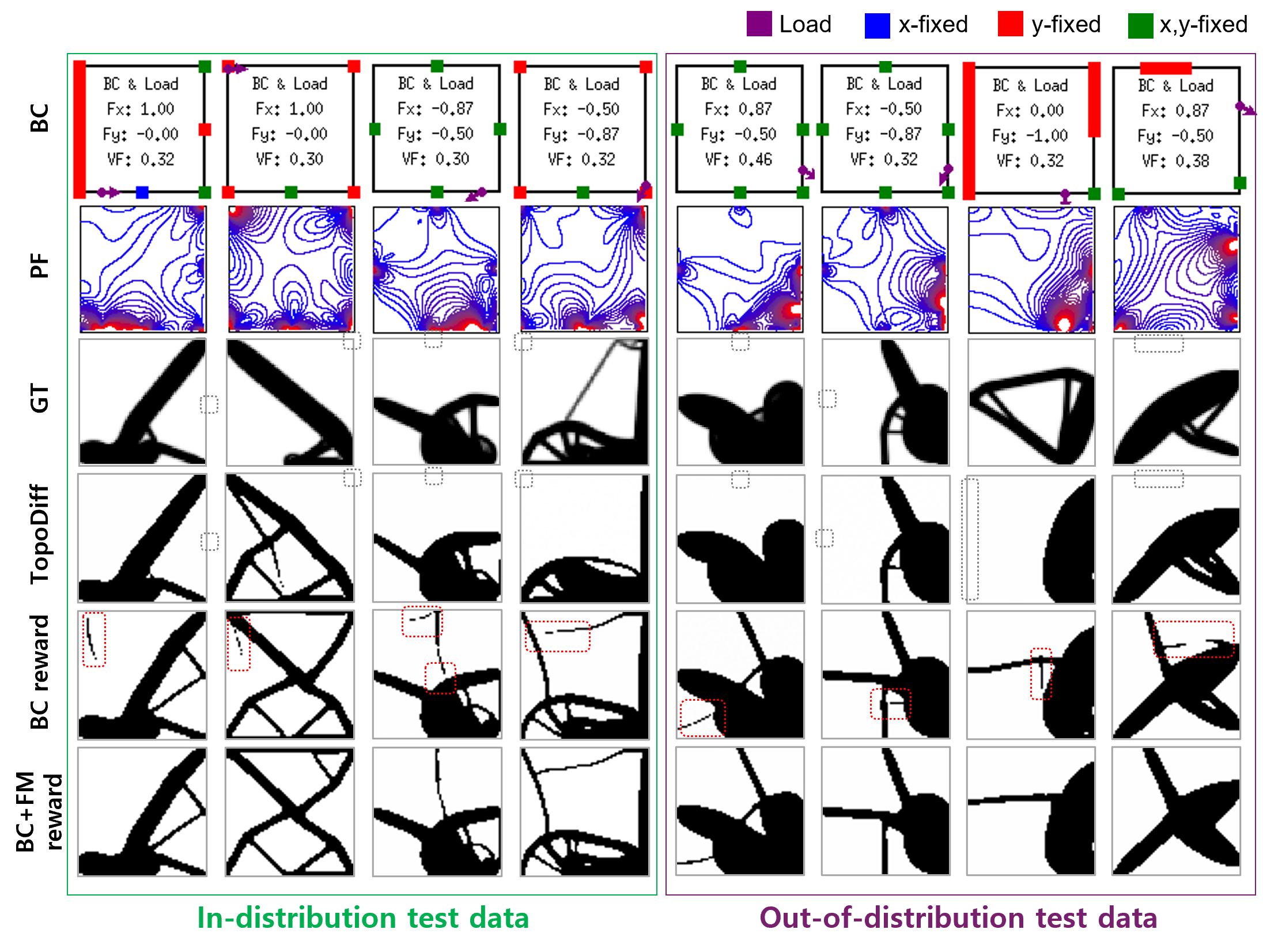}
\caption{
Case study illustrating the correction of two types of failure: violation of BCs and the presence of floating material. Each column corresponds to a distinct design condition, displaying: (top) prescribed boundary and loading conditions with volume fraction (VF); (second row) physical field visualization (e.g., strain energy); (third row) TopoDiff output exhibiting structural defects; (bottom rows) samples generated by our method using combined BC and FM reward gradients. 
In-distribution test cases (left) and out-of-distribution scenarios (right) demonstrate the generalization ability of our framework. Red boxes highlight violations that are mitigated in our outputs but persist in baseline results.
}
    \label{fig:case-joint}
\end{figure}

In this case study, we consider a set of combined scenarios where the TopoDiff model produces designs exhibiting both boundary violations and floating material which are two distinct types of structural infeasibility. Such dual failure modes are especially difficult to eliminate using a single surrogate or classifier.

We address these cases using our method with both reward models $R_{\psi_{\text{bc}}}$ and $R_{\psi_{\text{fm}}}$ applied jointly. As shown in Figure~\ref{fig:case-joint}, each row represents a different task condition (leftmost column), followed by the baseline TopoDiff output and multiple reward-guided samples (from left to right).

The baseline outputs exhibit clear defects: disconnected supports, floating fragments, and boundary detachment. After applying joint guidance with equal weighting $\lambda_{\text{bc}} = \lambda_{\text{fm}}$, the generated structures demonstrate significantly improved feasibility. Both BC and FM failures are suppressed, while preserving structural diversity and mechanical plausibility.

These findings confirm that jointly applying multiple reward models enables our sampler to resolve compound failures that cannot be addressed by any individual signal alone.

\subsection{Further improvement and quantitative evaluation}
As an extension, we further fine-tune the reward models by collecting additional human labels on new samples generated by our own reward-guided diffusion process. Rather than reusing model predictions, we present these guided samples to human evaluators for a second round of binary annotation. This iterative human-in-the-loop refinement exposes the reward model to more realistic and diverse failure modes, improving robustness and generalization in both in-distribution and out-of-distribution scenarios. 

In practice, we use an ensemble of the first-stage and second-stage reward models, combining their predictions via averaging. This ensembling strategy ensures that both early stage patterns and refined corrections contribute to the final guidance signal during sampling. 

\begin{figure}[H]
    \centering
    \includegraphics[width=\linewidth]{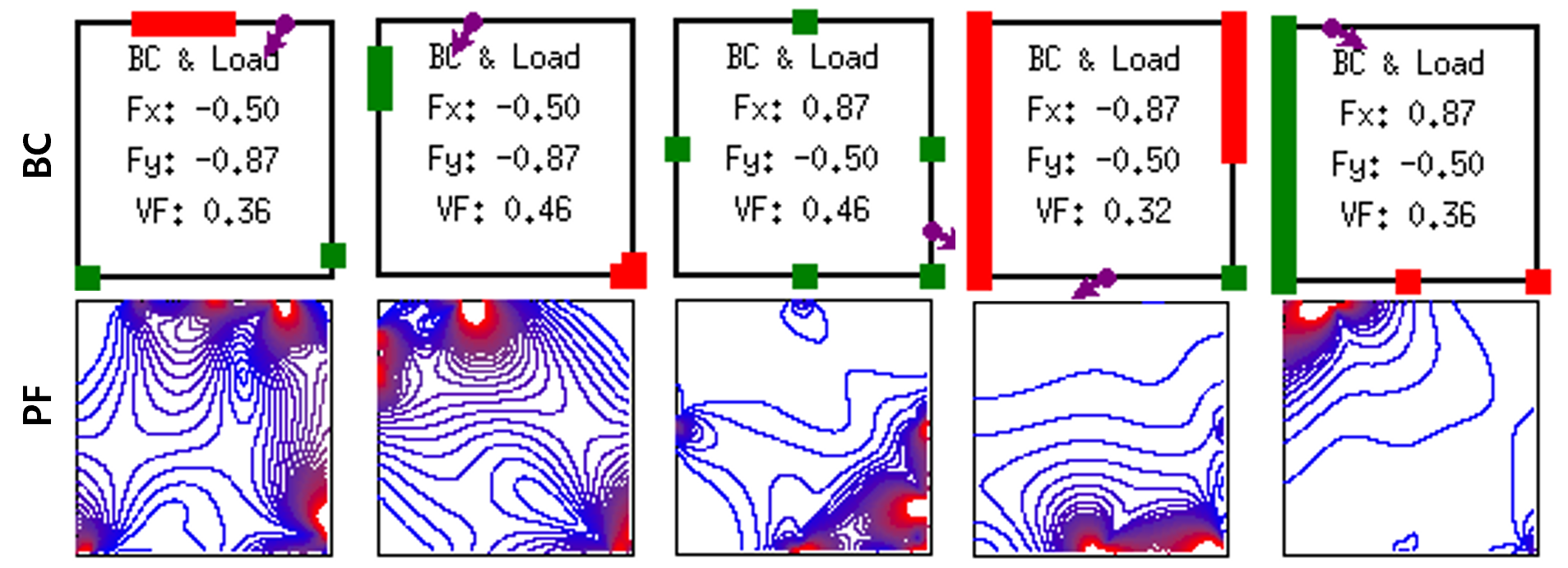}
    \caption{
Five randomly selected conditioning scenarios used in the quantitative evaluation. Each column shows a distinct test case, including prescribed boundary and load conditions, physical field inputs (e.g., strain energy), and sample outputs. These scenarios are not seen during training and reflect diverse loading configurations.
    }
    \label{fig:representative_scenario}
\end{figure}

We evaluate our method on 100 randomly selected conditioning inputs from the TopoDiff dataset, each specifying a unique combination of loads and boundary conditions (see Figure~\ref{fig:representative_scenario} for examples). For each input, we generate three samples using (1) the original TopoDiff model, (2) our method guided by the first-stage reward models, and (3) our method with reward models refined via second-stage feedback, resulting in a total of 900 samples.

We then compute the failure rate as the proportion of generated samples exhibiting the target violation type, as determined by a human evaluator. Each output is labeled as either \emph{valid} or \emph{invalid} for the designated failure type, consistent with the feedback criteria used for training the reward models.

\begin{figure}[H]
    \centering
    \includegraphics[width=\linewidth]{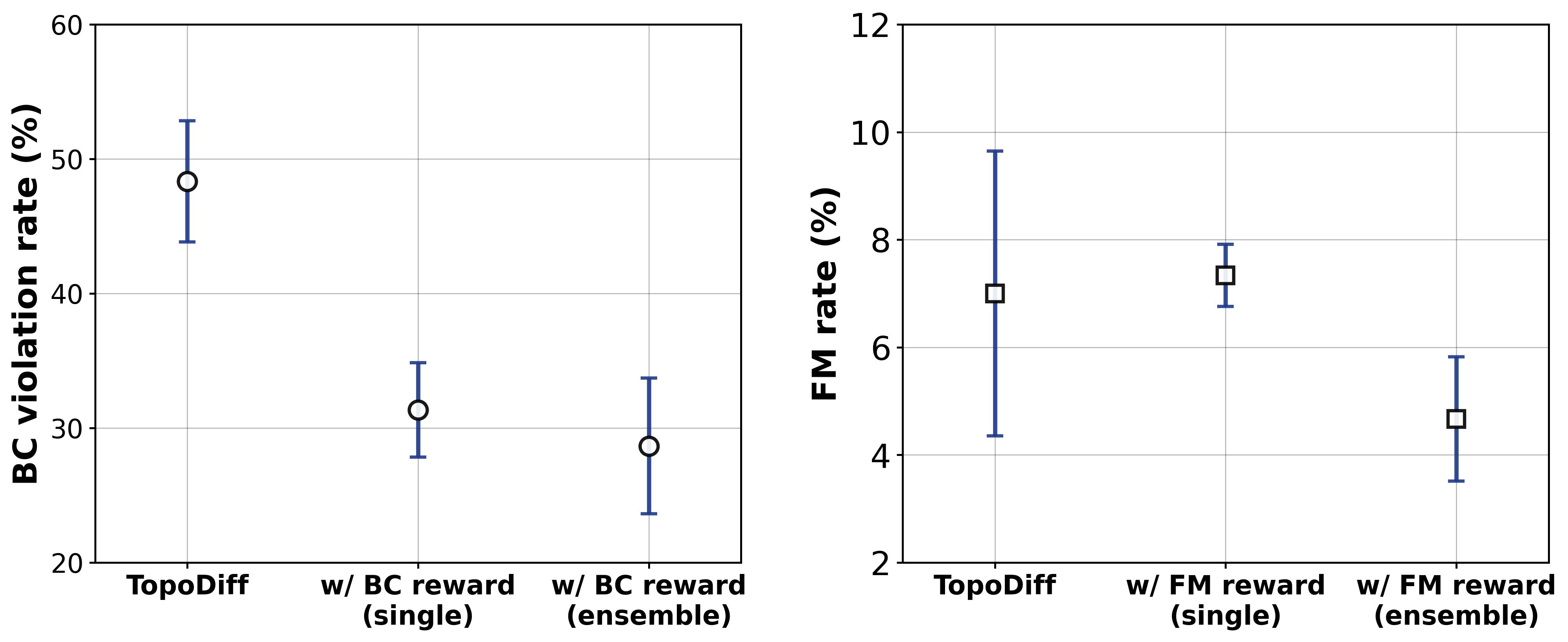}
    \caption{
        Failure rates across three evaluation categories. 
        Human-guided diffusion significantly reduces the frequency of invalid samples relative to the TopoDiff baseline. 
        Each bar reports the average failure rate 100 randomly selected conditioning inputs, with 3 samples per condition.
    }
    \label{fig:failure-rate}
\end{figure}

As shown in Figure~\ref{fig:failure-rate}, our method consistently reduces the rate of invalid generations across all failure types. In the BC category, the baseline model produces invalid outputs in 48.3\% $\pm$ 4.51, while the first-feedback guidance lowers this to 31.3\% $\pm$ 3.51, and the second-feedback stage further reduces it to 28.7\% $\pm$ 5.03.
Similarly, FM artifacts occur in {.0\% $\pm$ 2.65 of baseline samples, rise slightly to 7.3\% $\pm$ 0.58 after the first feedback, and then drop to 4.67\%$\pm$ 1.15 after the second feedback. 

Overall, these results demonstrate that reward-guided sampling improves structural validity in a consistent and measurable way. Notably, the second round of human feedback yields additional gains over the first-stage model, confirming the value of iterative refinement. Our method achieves these improvements without sacrificing diversity or requiring changes to the underlying diffusion model, making it practical and scalable for real-world generative design pipelines.



\section{Discussion}

Across all case studies and quantitative evaluations, we find that human-guided censored sampling reliably suppresses subtle geometric artifacts that are often missed by surrogate-based guidance alone. By incorporating lightweight reward models trained on a moderate amount of human-labeled data, our method effectively corrects structural failure modes such as boundary violations and disconnected components. The resulting designs remain physically valid and structurally compliant, with no noticeable loss in diversity or performance. Moreover, the framework is modular and easily extensible, allowing additional reward models targeting other constraint types such as symmetry or feature size to be integrated without modifying the diffusion backbone.

These results support the view that human feedback allows diffusion models to go beyond the limitations of fixed surrogates by capturing visually obvious design flaws that are difficult to encode analytically.

\paragraph{Limitations.}
Despite its advantages, our approach has several limitations. First, the reward models are only as good as the feedback they are trained on; sparse or biased labels may fail to capture certain pathologies, especially in complex boundary/load settings. Second, the reward models operate in isolation and assume that each violation type (e.g., BC or FM) is independent; in practice, these constraints may interact, and naive gradient combination can lead to suboptimal solutions or mode collapse. Third, our approach relies on binary labels and does not capture gradations of quality or user intent, which may be important in tasks involving aesthetics or trade-offs.

\paragraph{Future directions.}
Future work may explore more expressive feedback mechanisms, such as relative preference comparisons or continuous ratings. Incorporating uncertainty-aware reward models or active learning strategies could reduce annotation effort further. Additionally, extending this framework to 3D topology optimization, multi-material structures, or tasks involving safety-critical constraints (e.g., maximum stress) would significantly broaden its applicability. Finally, integrating reward guidance with PDE-informed learning~\cite{bastek2024physics} or differentiable simulation may offer hybrid approaches that combine human and physics priors systematically.

\bibliographystyle{IEEEtran}

\bibliography{ref}

\end{document}